\documentclass[journal]{IEEEtran}

\usepackage{amsmath,graphicx}
\usepackage[noadjust]{cite}
\usepackage[utf8]{inputenc}
\usepackage[T1]{fontenc}
\usepackage{float}
\usepackage{url}
\usepackage{grffile}
\usepackage{longtable}
\usepackage{wrapfig}
\usepackage{rotating}
\usepackage[normalem]{ulem}
\usepackage{textcomp}
\usepackage{amssymb}
\usepackage{capt-of}
\usepackage{leftindex}
\usepackage{hyperref}
\usepackage{amsthm}
\usepackage{mathtools}
\usepackage{centernot}
\usepackage{caption}
\usepackage{subcaption}
\usepackage{xcolor}
\usepackage[nolist]{acronym}
\usepackage{bm}

\newcommand{\mathbbkernmore}[1]{%
    \ooalign{$\mathbb{#1}$\cr$\mkern0.5mu\mathbb{#1}$\cr$\mkern1mu\mathbb{#1}$}%
}

\hypersetup{colorlinks=true}
\newcommand\df{\stackrel{\mathclap{\tiny\mbox{df.}}}{=}}
\newcommand{\sgens}[1]{\mathbb{R}^{#1}}
\newcommand{\sgen}[2]{\mathbb{R}^{#1\times #2}}

\newcommand{\cond}[0]{\operatorname{cond}}

\newtheorem{example}{Example}
\newtheorem{proposition}{Proposition}

\newtheorem{remark}{Remark}

\newtheorem{definition}{Definition}

\definecolor{revcolor}{rgb}{0,0,0.5}

\begin{acronym}[myacro]
  \acro{OTA}{Over-the-Air}
  \acro{FL}{federated learning}
  \acro{AWGN}{additive white Gaussian noise}
  \acro{ML}{machine learning}
\end{acronym}

\begin{document}

\title{Inverse Feasibility in Over-the-Air\\ Federated Learning}

\author{Tomasz Piotrowski, Rafail Ismayilov, Matthias Frey, Renato L.G. Cavalcante
\thanks{This work was in part supported by the National Centre for Research and
Development of Poland (NCBR) under grant EIG CONCERT-JAPAN/04/2021
and by the Federal Ministry of Education and Research of Germany under
grant 01DR21009 and the programme ``Souver\"an. Digital. Vernetzt.''
Joint project 6G-RIC, project identification numbers: 16KISK020K and
16KISK030.}
\thanks{Tomasz Piotrowski is with Nicolaus Copernicus University in Torun, Poland. (e-mail: tpiotrowski@umk.pl). Matthias Frey is with The University of Melbourne, Victoria, Australia. Renato L. G. Cavalcante and Rafail Ismayilov are with Fraunhofer Heinrich Hertz Institute, Berlin, Germany.}
\thanks{RI, MF, RLGC contributed equally to this work.}
\thanks{\textcopyright\ 2024 IEEE.  Personal use of this material is permitted.  Permission from IEEE must be obtained for all other uses, in any current or future media, including reprinting/republishing this material for advertising or promotional purposes, creating new collective works, for resale or redistribution to servers or lists, or reuse of any copyrighted component of this work in other works.}}
\maketitle

\begin{abstract}
We introduce the concept of inverse feasibility for linear forward models as a tool to enhance \ac{OTA} \ac{FL} algorithms. Inverse feasibility is defined as an upper bound on the condition number of the forward operator as a function of its parameters. We analyze an existing \ac{OTA} \ac{FL} model using this definition, identify areas for improvement, and propose a new \ac{OTA} \ac{FL} model. Numerical experiments illustrate the main implications of the theoretical results. The proposed framework, which is based on inverse problem theory, can potentially complement existing notions of security and privacy by providing additional desirable characteristics to networks.

\end{abstract}

%
\acresetall
\section{Introduction} \label{sec.introduction}

In recent years, wireless solutions have been increasingly applied to \ac{FL} because of their potential to distribute models efficiently, particularly in mobile environments. However, the physical properties of wireless channels present many challenges for \ac{FL}, including scalability, privacy, and security considerations. 

In particular, with traditional digital transmission schemes, interference caused by concurrent transmission of multiple users poses difficulties related to scalability, especially in dense networks operating in low frequency bands, which typically operate in interference limited regimes. To address this challenge, analogue \ac{OTA} transmission schemes have been proposed as a scalable alternative to traditional digital communication technologies in \ac{FL}. In \ac{OTA}, the superposition property of signals in the wireless medium is exploited to fuse the processes of communication and computation, which alleviates the need for complex scheduling and interference mitigation mechanisms. The scalability gains of OTA schemes are specially pronounced if a receiver has to compute a function of data distributed across a large network, which is a common scenario in \ac{FL} approaches. For example, during the learning stage of traditional \ac{FL} algorithms based on neural networks, a server may not be interested in obtaining gradients of the local cost functions of each node, but only in computing the average of all local gradients.

\ac{OTA} \ac{FL} may seem to improve security and privacy because individual nodes' data are not easily accessible (all signals are mixed in the wireless medium), and reconstructing information intended for the server depends on the propagation environment of malicious nodes (because \ac{OTA} is an analogue transmission scheme). However, we emphasize that there are no inherent guarantees of privacy and security. For this reason, researchers have been introducing many concepts related to privacy and security in \ac{OTA} \ac{FL} \cite{Geyer2017, Niknam2020, Wei2020, Zhang2020, Ma2020, Liu2020, frey2021towards-tit, frey2021towards-isit, Hu2022}. However, any security guarantee necessarily relies on additional assumptions, such as a computationally bounded adversary, a secret key only known to legitimate parties, or a bound on the attacker's channel strength. Therefore, a communication system that offers multiple security guarantees based on different assumptions can enhance resilience in the event that any individual assumption fails.

Against this backdrop, we propose a novel concept called inverse feasibility, which complements existing notions of security and privacy. Drawing on inverse problem theory, inverse feasibility leverages the fact that, during training, neural networks in \ac{OTA} \ac{FL} typically have to compute the average of a huge number of parameters distributed across nodes within a dense network. We assume that nodes apply possibly lossy compression schemes to local parameters prior to transmission, as proposed in \cite{Amiri2020, Amiri2020_2}, because of the limited bandwidth and high communication costs of wireless systems. As a result, the legitimate server needs to solve an ill-posed inverse problem to reconstruct the true average of parameters from the received compressed data. The core concept of inverse feasibility is that the difficulty of solving this inverse problem should be upper-bounded by a known function of the \ac{OTA} \ac{FL} model's parameters. We note that the proposed concept allows to evaluate this difficulty for the existing \ac{OTA} \ac{FL} models and also to introduce novel \ac{OTA} \ac{FL} models with improved inverse feasibility.
Building on this, the second main contribution of this study is the derivation of a new \ac{OTA} \ac{FL} compression scheme that 
enhances an existing \ac{OTA} \ac{FL} system proposed in \cite{Amiri2020, Amiri2020_2} with respect to our definition of inverse feasibility.


\section{System Model}
\label{sec.system_model}
In the text that follows, a vector or an operator with its domain and/or codomain involving all $M$ active nodes of an \ac{OTA} \ac{FL} scheme is denoted in bold letters such as $\bm{a}$ or $\bm{A}.$ A matrix written in blackboard bold font, such as $\mathbb{A}$, denotes a random matrix. 

To introduce the definition of inverse feasibility, we first need to describe an \ac{OTA} \ac{FL} model similar to that proposed in \cite{Amiri2020, Amiri2020_2}, which is the topic of this section. Briefly, at each iteration of the training stage of a machine learning model, a server tries to improve the parameters of the model by computing a global gradient, which is the average of gradients of local cost functions available to $M$ nodes in a network, via the analogue \ac{OTA} transmission scheme. By using this transmission scheme, the server acquires neither the local training sets nor the local gradients available to nodes. To describe the information obtained at the server, we denote the nodes' gradients of given functions (which are constructed from the nodes' datasets at a given iteration/time instant) at a differentiable point $\theta \in \mathbb{R}^d$ (the global model parameters to be learned collaboratively) as follows:
\begin{equation*}
\bm{g}(\theta)\df(g_1(\theta),\dots,g_M(\theta))\in\sgens{Md}.
\end{equation*}
To avoid cluttering equations, we abuse notation by omitting the dependency of gradients on $\theta$ in all equations that follow. For example, $\bm{g}=(g_1,\dots,g_M)\in\sgens{Md}$ should be understood as the vector $\bm{g}(\theta)=(g_1(\theta),\dots,g_M(\theta))\in\sgens{Md}$.

Prior to transmission, each node $m\in\{1,\ldots,M\}$ first computes a sparse version $g_m^{sp}\df S(g_m)\in\mathbb{R}^d$ of the local gradient $g_m \in\mathbb{R}^d$, where  $S\colon\mathbb{R}^d\to\mathbb{R}^d:x\mapsto x^{sp}$ and  
\begin{equation*}
	\left ( \forall i \in \left\{1,...,d \right\} \right ) [x^{sp}]_i = \left\{\begin{array}{ll}
		0, & \text{if} \hspace{4pt} |[x]_i|<\delta, \\
		\left [ x \right ]_i, & \text{otherwise}, \\
	\end{array}\right.
\end{equation*}
$\delta>0$ is the sparsification factor (a design parameter), and $[x]_i$ denotes the $i$th coordinate of the vector $x.$ 
In the second step, each node $m\in\{1,\ldots,M\}$ compresses the sparse gradient $g_m^{sp}$ by computing the matrix-vector multiplication $Ag_m^{sp}$, where $A\in\mathbb{R}^{s\times d}$ with $s<d$ is a compression matrix common to all nodes, and the entries of $A$ are samples of i.i.d. random variables distributed according to a normal distribution with zero mean and unity variance. 
With these operations, we assume that a sample of the signal received by the server with the analogue \ac{OTA} transmission scheme is given by
  \begin{equation} \label{eq.model_original_received}
    y=\frac{1}{M}\left(\sum_{m=1}^M h_m \sqrt{\alpha_m} A g_m^{sp}+\gamma\right),
  \end{equation}
  where $\sqrt{\alpha_m}>0$ represents a normalization constant applied by the transmitters to ensure that the transmitter power constraints are met, $h_m>0$ is a scalar that represents the slow, multiplicative channel fading, and $\gamma\in\mathbb{R}$ represents samples of \ac{AWGN}. The summation and application of fading and noise happen in the wireless channel, while $\sqrt{\alpha_m} A g_m^{sp}$ is the signal transmitted by node $m$.
The factor $1/M$ can be seen as a normalization applied by the receiver in post-processing.
We also assume that the channel fading coefficients $h_1, \dots, h_M$ and the normalization constants $\alpha_1, \dots, \alpha_M$ are known to the server. To simplify notation, we separate the linear and nonlinear components in~\eqref{eq.model_original_received} as follows. For the nonlinear component, we define the function 
  \begin{equation*}
    \bm{S}\colon\sgens{Md}\to\sgens{Md}:\bm{g}\mapsto \bm{g}^{sp},
  \end{equation*}
  where $\bm{g}^{sp}\df(g_1^{sp},\dots,g_M^{sp})\in\sgens{Md}$ is a vector containing all sparse gradients.


For the linear component, we define the matrix $\bm{L}(\underbar{p})\in\sgen{s}{Md}$, which is written in block form as
\begin{equation} \label{eq.model_original_decomposition}
\bm{L}(\underbar{p})\df\frac{1}{M}[h_1\sqrt{\alpha_1}A,\dots,h_M\sqrt{\alpha_M}A],
\end{equation}
where $\underbar{p}\in\underbar{P}$ and $\underbar{P}$ is the set of all possible parameters of the forward operator which is known at the server. As an example, for the \ac{OTA} \ac{FL} model \eqref{eq.model_original_received}, the set $\underbar{P}$ for the forward operator in~(\ref{eq.model_original_decomposition}) is comprised of tuples of the form $(M,d,s,A,h_1,\dots,h_M,\alpha_1,\dots,\alpha_M)$, with the respective domains for each parameter.



We emphasize in (\ref{eq.model_original_decomposition}) the dependency of the matrix $\bm{L}(\underbar{p})$ on the model parameters $\underbar{p}\in\underbar{P}$ for later use in the definition of inverse feasibility. With the above notation, the model \eqref{eq.model_original_received} can be equivalently written as:  
\begin{equation} \label{eq.fl_generic}
 y=\bm{f}  \left (\bm{g} \right )+\gamma/M,
\end{equation}  
where   
\begin{equation} \label{eq.f_decomp}
\bm{f}\colon\sgens{Md}\to\sgens{s}:\bm{g}\mapsto (\bm{L}(\underbar{p})\circ \bm{S}) \left (\bm{g}\right)=\bm{L}(\underbar{p})\bm{g}^{sp}.
\end{equation}  

\section{Inverse Feasibility in Over-the-Air Federated Learning: Definition and Examples}\label{sec.def_inv_feas}

\subsection{Definition of Inverse Feasibility} \label{def}

In the federated learning task under consideration, the objective of the server is to reconstruct the average of gradients, defined as
\begin{equation} \label{avg}
\sgens{d}\ni\bar{g}^{sp} \df \frac{1}{M}\sum_{m=1}^Mg_m^{sp}=\frac{1}{M}\underbrace{[I_d,\dots,I_d]}_M\bm{g}^{sp},
\end{equation}
 by solving an inverse problem 
from the compressed received signal $y$ in (\refeq{eq.model_original_received}). Following a standard assumption in the literature, we assume that  $\bar{g}^{sp}$ is sparse (see, e.g., \cite[Assumption 3]{Amiri2020}), which is often justified by the empirical evidence that, in some learning tasks, the sparsity pattern of $g_m^{sp}$ for $m=1,\ldots,M$ in $\bm{g}^{sp}$ is common across the $M$ active nodes in the network. 

In view of (\ref{eq.fl_generic})-(\ref{eq.f_decomp}), and recalling that the condition number of a linear operator is a classical tool for assessing the difficulty of solving inverse problems (see, e.g., \cite{Engl1996,Golub1996} and references therein), we posit that a desirable condition for the estimation of any function of $\bm{g}^{sp}$, such as the arithmetic mean in (\ref{avg}), is that the condition number of the linear operator $\bm{L}(\underbar{p})$ should be close to one. The concept of inverse feasibility we propose is based on this assumption. 
Formally, we define inverse feasibility as follows:
\begin{definition}[inverse feasibility] \label{def.inverse_feasibility}
We call an \ac{FL} model of the form in \eqref{eq.fl_generic} at least $F$-inverse feasible for a given function $F\colon\underbar{P}\to[1,\infty]$ if
\begin{equation} \label{inv}
(\forall\underbar{p}\in\underbar{P})~\cond(\mathbbkernmore{L}(\underbar{p}))\leq F(\underbar{p}),\quad a.s.,
\end{equation}
where the condition number of $\mathbbkernmore{L}(\underbar{p})$ is a random variable defined in the following standard way:
\begin{equation}
\cond \left (\mathbbkernmore{L}(\underbar{p}) \right )\df\sigma_1 \left (\mathbbkernmore{L}(\underbar{p}) \right )/\sigma_s \left (\mathbbkernmore{L}(\underbar{p}) \right )\geq 1,
\end{equation}
where $\sigma_1 \left (\mathbbkernmore{L}(\underbar{p}) \right )$ and $\sigma_s \left (\mathbbkernmore{L}(\underbar{p}) \right )$ are the largest and the smallest singular values of $\mathbbkernmore{L}(\underbar{p})$, respectively. (We assume that, in practical settings, we have $\sigma_s \left (\mathbbkernmore{L}(\underbar{p}) \right )>0$ almost surely.)
\end{definition}


\begin{remark}
The concept of inverse feasibility introduced in Definition \ref{def.inverse_feasibility} enables us to provide weaker notions of inverse feasibility, such as $F$-inverse feasibility in the mean sense,  
\begin{equation} \label{inv_e}
(\forall\underbar{p}\in\underbar{P})~\mathbb{E}[\cond(\mathbbkernmore{L}(\underbar{p}))]\leq F(\underbar{p}),
\end{equation}
and $F$-inverse feasibility with probability of at least $r\in[0,1]$,
\begin{equation} \label{inv_p}
(\forall\underbar{p}\in\underbar{P})~\mathbb{P}[\cond(\mathbbkernmore{L}(\underbar{p}))\leq F(\underbar{p})]\geq r. 
\end{equation}
We will use the concept of $F$-inverse feasibility with probability for the model to be proposed in Example \ref{exmpl.model_legitimate_server} in Section \ref{main} below. For brevity, the concept of $F$-inverse feasibility in the mean sense is not used in this letter. 
\end{remark}


Ideally, for well-behaved \ac{OTA} \ac{FL} model, there should exist a function $F$ taking values close to one for every possible $\underbar{p}\in\underbar{P}$ at the server. 
We also emphasize that the concept of inverse feasibility provides us with a means of quantifying a desirable property of \ac{OTA} \ac{FL} models, but it does not replace existing concepts of security and privacy in wireless systems. 

\vspace{-\baselineskip}
\subsection{Inverse Feasibility of \ac{OTA} \ac{FL} Models} \label{main}

We now proceed to analyze the \ac{OTA} \ac{FL} model in \cite{Amiri2020, Amiri2020_2} with respect to Definition~\ref{def.inverse_feasibility}, and later we propose enhancements. 

\begin{proposition} \label{p1}
The model \eqref{eq.model_original_received} satisfies $F(\underbar{p})\df\cond(A)\equiv\cond(\bm{L(p)}).$ 
\end{proposition}
\proof See the supplementary material.

We recall that the set $\underbar{P}$ depends on all model parameters available at the server, but Proposition~\ref{p1} shows that the inverse feasibility of the \ac{OTA} \ac{FL} model (\ref{eq.model_original_received}) depends only on the condition number of the compression matrix $A.$ This result suggests that we may improve the inverse feasibility of the \ac{OTA} \ac{FL} model \eqref{eq.model_original_received} by replacing the deterministic compression matrix~$A$ with a compression matrix $W^\star\in\sgen{s}{d}$ optimized to have a small condition number, so that the forward operator becomes in this case
\begin{equation} \label{eq.model_original_decomposition_W}
\bm{L}(\underbar{p})\df\frac{1}{M}[h_1\sqrt{\alpha_1}W^\star,\dots,h_M\sqrt{\alpha_M}W^\star].
\end{equation}
This matrix $W^\star$ can be obtained, for example, as the solution to an optimization problem designed to minimize the mutual coherence of the columns of $W^\star$ \cite{Abolghasemi2010}. The resulting \ac{OTA} \ac{FL} model is thus a minor variation on the model \eqref{eq.model_original_received}:
\begin{equation} \label{model_W}
y=\frac{1}{M}\left(\sum_{m=1}^M h_m \sqrt{\alpha_m} W^\star g_m^{sp}(\theta)+\gamma\right).
\end{equation}
Proceeding analogously as in the proof of Proposition \ref{p1}, we can deduce that $\cond(W^\star)\equiv\cond(\bm{L(p)}).$ From a practical standpoint, the optimization problem for constructing $W^\star$ can be solved at the server, and then the resulting matrix can be broadcast to all active nodes. Later, in Section \ref{sec.numerical_evaluation}, we show the performance of this approach via simulations.

As an alternative to the approach described above, nodes can independently take samples of random matrices with means given by a common matrix known to the server and all nodes, for example, $A$ or $W^\star.$ The advantage of this alternative, which is illustrated in Example~\ref{exmpl.model_legitimate_server} below, is that each user has its own local compression matrix, which is not transmitted across the network.

\begin{example} \label{exmpl.model_legitimate_server}
Consider the following \ac{OTA} \ac{FL} model:
  \begin{equation} \label{eq.model_legitimate_server_received}
    y=\frac{1}{M}\left(\sum_{m=1}^M h_m \sqrt{\alpha_m} \mathbb{B}_m g_m^{sp}+\gamma\right),
\end{equation}
  where $\mathbb{B}_m=X+\mathbb{G}_m$, $X\in\sgen{s}{d}$ is an arbitrary fixed matrix (e.g., $X=A$ or $X=W^\star$) common to all nodes, and $\mathbb{G}_m$ is a random matrix with realizations in $\sgen{s}{d}$ and all of its entries identically and independently distributed according to $\mathcal{N}(0,\sigma^2).$ In this case, $\underbar{P}$ consists of tuples of the form
\begin{equation*}  
(M,d,s,X,\mathbb{G}_1,\dots,\mathbb{G}_M,h_1,\dots,h_M,\alpha_1,\dots,\alpha_M),
\end{equation*}  
with the following block form for the matrix $\mathbbkernmore{L}(\underbar{p})$:
  \begin{equation} \label{eq.model_legitimate_server_decomposition}
    \mathbbkernmore{L}(\underbar{p})=\frac{1}{M}(h_1\sqrt{\alpha_1}\mathbb{B}_1,\dots,h_M\sqrt{\alpha_M}\mathbb{B}_M).
  \end{equation}
\end{example}
We note that the model (\ref{eq.model_legitimate_server_received}) is similar to privacy-enhancing models such as those in \cite{Wei2020}, because, for each node $m\in M$, the term $\mathbb{G}_mg_m^{sp}$ introduces noise perturbation on the transmitted gradients.

The next proposition establishes the inverse feasibility of the model in Example~\ref{exmpl.model_legitimate_server} in probabilistic terms. Later, in Section \ref{pc}, we evaluate this model in a scenario where power control is applied at user nodes.

\begin{proposition} \label{p3}
For every $t\geq 0$, the \ac{OTA} \ac{FL} model in Example~\ref{exmpl.model_legitimate_server} is $F_t$-inverse feasible with probability of at least $1-2e^{\frac{-t^2}{2}}$; i.e.,
\begin{equation} \label{inv_p2}
(\forall\underbar{p}\in\underbar{P})\ \mathbb{P}[\cond(\mathbbkernmore{L}(\underbar{p}))\leq F_t(\underbar{p})]\geq 1-2e^{\frac{-t^2}{2}},
\end{equation}
where
\begin{equation} \label{FT}
F_t(\underbar{p})\df  \left\{
  \begin{array}{ll}
    \frac{\sqrt{c}\sigma_1(X)+x_t}{\sqrt{c}\sigma_s(X)-x_t}, & \sqrt{c}\sigma_s(X)-x_t>0,\\
    \infty & otherwise.\\
  \end{array}\right.
\end{equation}
and where
\begin{itemize}
\item $\sigma_1(X)$, $\sigma_s(X)$ are the largest and the smallest singular value of $X$, respectively;
\item $c_m\df h_m\sqrt{\alpha_m}$ for $m=1,\dots,M$;
\item $c_{\max}\df\max_{m\in\{1,\dots,M\}}c_m$;
\item $c\df\sum_{m=1}^Mc_m^2$; and
\item $x_t\df\sigma c_{\max}(\sqrt{Md}+\sqrt{s}+t)$ for $t\geq 0$,
where $\sigma^2>0$ is the variance of random matrices $\mathbb{G}_m$ in model~(\ref{eq.model_legitimate_server_received}).
\end{itemize}
\end{proposition}
\proof See supplementary material.

\begin{remark} \label{prob_1}
Taking into account the typical floating-point precision of modern computers, we verify that the probability in (\ref{inv_p2}) is practically equal to one for $t\geq 10.$
\end{remark}

\vspace{-\baselineskip}
\section{Power control} \label{pc}
\subsection{Evaluation of Proposition \ref{p3}}
If users are able to apply power control to ensure that the product of the channel fading coefficient $h_m$ and the power scaling coefficient $\sqrt{\alpha_m}$ is similar for every user $m$ (e.g., $h_m\sqrt{\alpha_m}=1$ for every user $m$), then Proposition \ref{p3} offers a specially simple interpretation of the bound in (\ref{inv_p2}). 

\begin{remark} \label{sigma}
Let $c_{\min}\df\min_{m\in\{1,\dots,M\}}c_m$ and let $c_{\max}=c_{\min}=1$, in which case $c=M.$ The function $F_t$ in (\ref{FT}) yields meaningful bounds ($\infty~>~F_t(\underbar{p})\geq~1)$ if $\sqrt{c}\sigma_s(X)-x_t>0$, which, with power control, results in the constraint on the standard deviation of the entries of the perturbation matrices $\mathbb{G}_m$ of the form $\sigma<\frac{\sqrt{M}\sigma_s(X)}{\sqrt{Md}+\sqrt{s}+t}.$  
\end{remark}  




To put the above results in a numerical perspective, we consider in Section~\ref{sec.numerical_evaluation} a numerical setup with $t=10$, $d=7850$, $s=3925$, $M\geq 1e+3$, $\sigma_1(A)\approx 151$, and $\sigma_s(A)\approx 26$ (so that $\cond(A)\approx 5.8$), yielding $\sigma<0.28$ in Remark \ref{sigma}. Moreover, $\sigma_1(W^\star)\approx 141.47$, $\sigma_s(W^\star)\approx 141.42$ (so that $\cond(W^\star)\approx 1$), with $\sigma<1.55$ in Remark \ref{sigma}. In this case,
\begin{itemize}
\item $F_{10}(M=1e+3,\dots,\sigma_1(A),\sigma_s(A))<9.43$,
\item $F_{10}(M=1e+4,\dots,\sigma_1(A),\sigma_s(A))<9.33$,
\item $F_{10}(M=1e+3,\dots,\sigma_1(W^\star),\sigma_s(W^\star))<1.14$,
\item $F_{10}(M=1e+4,\dots,\sigma_1(W^\star),\sigma_s(W^\star))<1.14.$
\end{itemize}
In particular, Proposition \ref{p3} reveals that, by introducing the local perturbations $\mathbb{G}_m$ to the model (\ref{eq.model_legitimate_server_received}), we increase only marginally the condition number of the forward operator $\mathbbkernmore{L}(\underbar{p})$ in (\ref{eq.model_legitimate_server_decomposition}). Hence, assuming power control, we  anticipate that the reconstruction performance of the (averaged) gradient is only minimally impacted by adding local noise $\mathbb{G}_m$ to the global compression matrix $X$ at each node $m$. Moreover,  using $X=W^\star$ as the common matrix known to the server in the model (\ref{eq.model_legitimate_server_received}), we achieve the lower bound on the condition number $\cond(\mathbbkernmore{L}(\underbar{p}))$ of the forward operator $\mathbbkernmore{L}(\underbar{p})$ and also the higher allowable variance $\sigma^2$ of the perturbation matrices $\mathbb{G}_m$, compared with $X=A.$

\subsection{Parameter Estimation} \label{pe}

We now present a method for estimating the average of sparse gradients $\bar{g}^{sp}$ in (\ref{avg}) using the compressed signal $y$ received at the server. Our objective is to demonstrate that Definition~\ref{def.inverse_feasibility}, which is independent of any specific algorithm, offers valuable insights into the difficulty associated with reconstructing $\bar{g}^{sp}$ from $y$ using classical algorithms in the literature.

In more detail, the estimation of $\bar{g}^{sp}$ in \eqref{eq.model_original_received}, (\ref{model_W}), \eqref{eq.model_legitimate_server_received}, can be posed as the following compressed sensing recovery problem, which is also known as the least absolute shrinkage and selection operator (lasso) regression:
\begin{equation} \label{eq.parameter_estimation}
\underset{\widehat{\bar{g}}^{sp}}{\text{minimize }} \left [ \dfrac{1}{2} \left\| y - Y \widehat{\bar{g}}^{sp} \right\|_2^2 + \beta \left\| \widehat{\bar{g}}^{sp} \right\|_1\right ],
\end{equation}
where $Y$ is the average of the local compression matrices used by the nodes, and $\beta>0$ is a design parameter used to control the sparsity of the solution, which is the estimate of $\bar{g}^{sp}$ in~(\ref{avg}). Let $1_M\in\sgen{M}{M}$ be the matrix of ones. Then, the actual forward operator of the form (\ref{eq.f_decomp}) used in models (\ref{eq.model_original_received}), (\ref{model_W}), (\ref{eq.model_legitimate_server_received}), is replaced for the lasso regression with the approximating (averaged) forward operator $\bm{g}\mapsto \bm{L}(\underbar{p})(1_M\otimes I_d)(\bm{g}^{sp}/M)$, which assumes in particular that the server has access to the arithmetic mean of users' forward operators.

In more detail, in \eqref{eq.model_original_received} or (\ref{model_W}), the optimization algorithm aims to recover $\bar{g}^{sp}$ in (\ref{avg}) using an arithmetic mean of the user blocks of the forward operator $\bm{L}(\underbar{p})$ in (\ref{eq.model_original_decomposition}) or (\ref{eq.model_original_decomposition_W}), respectively:
\begin{equation} \label{agg1}
Y=\bm{L}(\underbar{p})\underbrace{[I_d,\dots,I_d]^t}_M,
\end{equation}
which simplifies to $Y=A$ or $Y=W^\star$ for the case where power control is applied. On the other hand, in model \eqref{eq.model_legitimate_server_received}, the lasso algorithm attempts to use the following random variable, which is a function of the forward operator $\mathbbkernmore{L}(\underbar{p})$ in (\ref{eq.model_legitimate_server_decomposition}):
\begin{equation} \label{agg2}
\mathbbkernmore{Y}=\mathbbkernmore{L}(\underbar{p})\underbrace{[I_d,\dots,I_d]^t}_M.
\end{equation}
However, $\mathbbkernmore{Y}$ cannot be evaluated at the server owing to the user-dependent compression matrices $\mathbb{B}_m=X+\mathbb{G}_m$ in model~\eqref{eq.model_legitimate_server_received}. Nevertheless, the server has access to $\mathbb{E}[\mathbbkernmore{Y}] = X$, where $X=A$ or $X=W^\star$, which is the compression matrix we propose to use in (\ref{eq.parameter_estimation}) to estimate $\bar{g}^{sp}$ from the received signal $y$ for model \eqref{eq.model_legitimate_server_received}. We note that the variance of the entries of $\mathbbkernmore{Y}$ is $\sigma^2/M$, thus the error caused by using $X$ in place of $\mathbbkernmore{Y}$ decreases with~$M.$ Indeed, we can provide an upper bound for $\cond(\mathbbkernmore{Y})$ in the special case of reconstructing the averaged gradient using lasso regression. The details can be found in the supplementary material.

\subsection{Numerical Evaluation} \label{sec.numerical_evaluation}

\begin{figure}[t] \includegraphics[clip,width=\columnwidth]{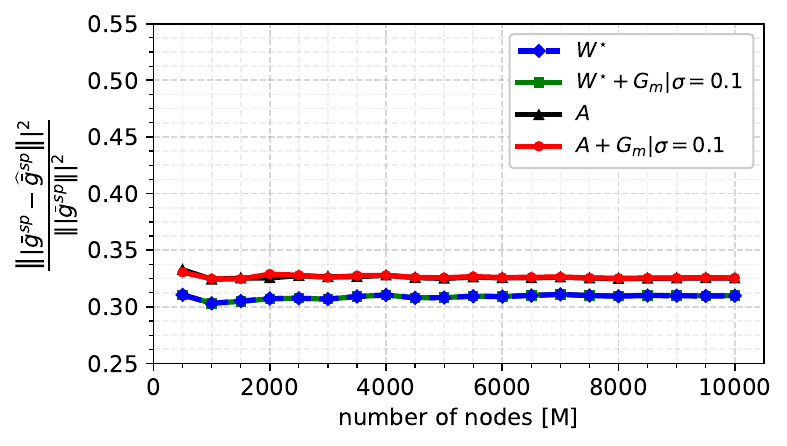}
\caption{Reconstruction error of \ac{OTA} \ac{FL} models against the number of active nodes $M.$}
\label{fig.sim_results}
\end{figure}

In our experiments, we use the MNIST dataset \cite{Lecun1998}, and we train a single-layer neural network similar to the one proposed in \cite{Amiri2020_2}, which consists of $d = 7850$ parameters. This neural network is trained using the ADAM optimizer with a learning rate of 0.001. The training process includes 100 epochs, where each node trains its model with 1000 samples and a batch size of 200. To predict the parameters, we use the \texttt{sklearn.linear\_model.Lasso} tool \cite{sklearn2022}. The compression ratio is $d/s=2$, with $\beta$ optimized separately for each model and $\gamma \sim \mathcal{N} (0, \rho^2)$ with $\rho = 0.01.$ We recall that in this Section \ref{pc} we assume that users apply power control to ensure that the product of the channel fading coefficient $h_m$ and the power scaling coefficient $\sqrt{\alpha_m}$ is normalized to $h_m\sqrt{\alpha_m}=1$ for every user $m.$ In this setup, we were able to confirm numerically the findings discussed in Section \ref{pc}. More specifically, the performance deterioration caused by introduction of perturbation matrices $\mathbb{G}_m$ in model~(\ref{eq.model_legitimate_server_received}) was insignificant (see Fig.~\ref{fig.sim_results}). The error floor for the model employing the compression matrix $W^\star$ is caused by the limitations of lasso regression and the additive noise $\gamma$ present in estimation.
\acresetall

\section{Conclusion and final remarks}
We have introduced the concept of inverse feasibility for general linear forward models, and we demonstrated its usefulness in OTA FL applications. The notion of inverse feasibility is rooted in the idea that, in inverse problems, linear forward operators should have a low condition number to ensure good reconstruction performance. We illustrated the validity of this assumption in an OTA FL problem focused on reconstructing the average of gradients. One potential application is to provide FL systems with additional guarantees against attacks. More precisely, by designing FL systems to guarantee that eavesdroppers have a high condition number of the forward operator, we expect to equip the system with an additional level of protection that complements existing notions of security and privacy.  Finally, our results have the potential to be extended to more general nonlinear forward models, opening up opportunities for further research in this area.



\section{Supplementary material}
\subsection{Proof of Proposition 1}
For brevity, we denote $\bm{L(p)}\mapsto\bm{L}.$ We note that the condition number does not depend on rescaling a given matrix by a fixed scalar, so it is sufficient to investigate the condition number of
$
\bm{L}'=(h_1\sqrt{\alpha_1}A,\dots,h_M\sqrt{\alpha_M}A).
$
Let $\sigma_i(X)$ and $\lambda_i(Y)$ denote the $i$-th singular value and eigenvalue of a given matrix $X$ and a symmetric matrix $Y$, respectively, both organized in nonincreasing order. Then
\begin{multline*}
\sigma_i^2(\bm{L}')=\lambda_i((\bm{L}')(\bm{L}')^t)=\\\lambda_i(\sum_{m=1}^Mh_m^2\alpha_mAA^t)=\sigma_i^2(A)\sum_{m=1}^Mh_m^2\alpha_m.
\end{multline*}
Let $c\df\sum_{m=1}^Mh_m^2\alpha_m>0.$ Then $\sigma_i(\bm{L}')=\sqrt{c}\sigma_i(A)$ for $i=1,2,\dots,s.$ Therefore,
\begin{multline*}
\forall M\in\mathbb{N}\quad \cond(\bm{L})=\cond(\bm{L}')=\\\frac{\sigma_1(\bm{L}')}{\sigma_s(\bm{L}')}=\frac{\sigma_1(A)}{\sigma_s(A)}=\cond(A).\ \blacksquare
\end{multline*}

\subsection{Proof of Proposition 2}
For brevity, we denote $\mathbbkernmore{L}(\bm{p})\mapsto\mathbbkernmore{L}.$ Let us consider the condition number of $\mathbbkernmore{L}$ in (12), or, equivalently, the condition number of
$
\mathbbkernmore{L}'=(c_1\mathbb{B}_1,\dots,c_m\mathbb{B}_M).
$
From \cite[Theorem 3.3.16]{Horn1991} and definition of $\mathbb{B}_m$, we have
\begin{equation} \label{s1}
\sigma_1(\mathbbkernmore{L}')\leq\sigma_1(c_1X,\dots,c_MX)+\sigma_1(c_1\mathbb{G}_1,\dots,c_M\mathbb{G}_M).
\end{equation}
From the proof of Proposition 1 we conclude that
\begin{equation} \label{odp1}
\sigma_i(c_1X,\dots,c_MX)=\sqrt{c}\sigma_i(X),\quad i=1,\dots,s.
\end{equation}
Furthermore, 
\begin{multline} \label{Gauss}
\sigma_1(c_1\mathbb{G}_1,\dots,c_M\mathbb{G}_M)\leq c_{\max}\sigma_1(\mathbb{G}_1,\dots,\mathbb{G}_M)=\\c_{\max}\sigma_1(\sigma \mathbb{G}'_1,\dots,\sigma \mathbb{G}'_M)=\sigma c_{\max}\sigma_1(\mathbb{G}'_1,\dots,\mathbb{G}'_M),
\end{multline}
where $\mathbb{G}'_m$ has all entries i.i.d. $\mathcal{N}(0,1)$, $m=1,\dots,M.$ Let $\mathbbkernmore{G}'\df(\mathbb{G}'_1,\dots,\mathbb{G}'_M).$ Then, from \cite[Ineq. (2.3)]{Rudelson2010}, one has for $t\geq 0$ that
\begin{equation} \label{Rudy}
\mathbb{P}(\sigma_1(\mathbbkernmore{G}')\leq\sqrt{Md}+\sqrt{s}+t)\geq 1-2e^{\frac{-t^2}{2}}.
\end{equation}
From (\ref{s1})-(\ref{Rudy}) we obtain therefore that
\begin{equation} \label{nom_new}
\sigma_1(\mathbbkernmore{L}')\leq\sqrt{c}\sigma_1(X)+\sigma c_{\max}(\sqrt{Md}+\sqrt{s}+t),
\end{equation}  
with probability at least $1-2e^{\frac{-t^2}{2}}.$
Similarly, from \cite[Theorem 3.3.16]{Horn1991} (see also \cite{Loyka2015}) and (\ref{odp1})-(\ref{Gauss}) we also have that
\begin{equation*}
\sigma_s(\mathbbkernmore{L}')\geq\sqrt{c}\sigma_s(X)-\sigma c_{\max}\sigma_1(\mathbbkernmore{G'}),
\end{equation*}
thus, from (\ref{Rudy}), one has
\begin{equation}
\sigma_s(\mathbbkernmore{L}')\geq\sqrt{c}\sigma_s(X)-\sigma c_{\max}(\sqrt{Md}+\sqrt{s}+t),
\end{equation}
with probability at least $1-2e^{\frac{-t^2}{2}}.$ Therefore, if $\sqrt{c}\sigma_s(X)-x_t>0$,
we conclude that
\begin{equation*}
\cond(\mathbbkernmore{L})\leq\frac{\sqrt{c}\sigma_1(X)+x_t}{\sqrt{c}\sigma_s(X)-x_t}\geq 1,
\end{equation*}
with probability at least $1-2e^{\frac{-t^2}{2}}.\ \blacksquare$

\subsection{Condition Number of the Operator in Eq. (18)}
We recall that $c_{\max}=c_{\min}=1$ by assumption. Thus, we have 
\begin{equation*}
\mathbbkernmore{Y}=\frac{1}{M}\sum_{m=1}^M\mathbb{B}_m=X+\frac{1}{M}\sum_{m=1}^M\mathbb{G}_m,
\end{equation*}  
and hence, from \cite[Theorem 3.3.16]{Horn1991}, one has
\begin{equation*}
\sigma_1(\mathbbkernmore{Y})\leq\sigma_1(X)+\frac{1}{\sqrt{M}}\sigma\sigma_1(\mathbb{G}),
\end{equation*}
where the entries of the matrix $\mathbb{G}$ are all i.i.d.  $\mathcal{N}(0,1).$ Furthermore, from \cite[Ineq. (2.3)]{Rudelson2010} one has for $t\geq 0$ that
\begin{equation} \label{rudy2}
\mathbb{P}(\sigma_1(\mathbb{G})\leq\sqrt{d}+\sqrt{s}+t)\geq 1-2e^{\frac{-t^2}{2}}.
\end{equation}
Thus,
\begin{equation*}
\sqrt{M}\sigma_1(\mathbbkernmore{Y})\leq\sqrt{M}\sigma_1(X)+\sigma(\sqrt{d}+\sqrt{s}+t),
\end{equation*}
with probability at least $1-2e^{\frac{-t^2}{2}}.$ Similarly, from \cite[Theorem 3.3.16]{Horn1991} (see also \cite{Loyka2015}) we have
\begin{equation*}
\sigma_s(\mathbbkernmore{Y})\geq\sigma_s(X)-\frac{1}{\sqrt{M}}\sigma\sigma_1(\mathbb{G}),
\end{equation*}
thus, from (\ref{rudy2}), one has
\begin{equation*}
\sqrt{M}\sigma_s(\mathbbkernmore{Y})\geq\sqrt{M}\sigma_s(X)-\sigma(\sqrt{d}+\sqrt{s}+t),
\end{equation*}
with probability at least $1-2e^{\frac{-t^2}{2}}.$ Therefore, as long as $\sqrt{M}\sigma_s(X)>\sigma(\sqrt{d}+\sqrt{s}+t)$,
\begin{equation} \label{prob2}
\cond(\mathbbkernmore{Y})\leq\frac{\sqrt{M}\sigma_1(X)+\sigma(\sqrt{d}+\sqrt{s}+t)}{\sqrt{M}\sigma_s(X)-\sigma(\sqrt{d}+\sqrt{s}+t)},
\end{equation}
with probability at least $1-2e^{\frac{-t^2}{2}}.$ We note that we obtained an upper bound for $\cond(\mathbbkernmore{Y})$ in a very similar form to the assertion of Proposition 2.

\bibliographystyle{IEEEbib}
\bibliography{strings,refs}

\begin{thebibliography}{10}

\bibitem{Geyer2017}
R.~C. Geyer, T.~Klein, and M.~Nabi,
\newblock ``Differentially private federated learning: A client level
  perspective,'' 2017,
\newblock arXiv.

\bibitem{Niknam2020}
S.~Niknam, H.~S. Dhillon, and Reed J.~H,
\newblock ``Federated learning for wireless communications: Motivation,
  opportunities, and challenges,''
\newblock {\em IEEE Communications Magazine}, vol. 58, no. 6, pp. 46--51, 2020.

\bibitem{Wei2020}
K.~Wei, J.~Li, M.~Ding, C.~Ma, H.~H. Yang, F.~Farokhi, S.~Jin, T.~Q.~S. Quek,
  and H.~V. Poor,
\newblock ``Federated learning with differential privacy: Algorithms and
  performance analysis,''
\newblock {\em IEEE Transactions on Information Forensics and Security}, vol.
  15, pp. 3454--3469, 2020.

\bibitem{Zhang2020}
C.~Zhang, S.~Li, J.~Xia, W.~Wang, F.~Yan, and Y.~Liu,
\newblock ``{BatchCrypt}: Efficient homomorphic encryption for {Cross-Silo}
  federated learning,''
\newblock in {\em 2020 USENIX Annual Technical Conference (USENIX ATC 20)},
  2020.

\bibitem{Ma2020}
C.~Ma, J.~Li, M.~Ding, H.~H. Yang, F.~Shu, T.~Q.~S. Quek, and H.~V. Poor,
\newblock ``On safeguarding privacy and security in the framework of federated
  learning,''
\newblock {\em IEEE Network}, vol. 34, no. 4, pp. 242--248, 2020.

\bibitem{Liu2020}
Y.~Liu, J.~Peng, J.~Kang, A.~M. Iliyasu, D.~Niyato, and A.~A.~Abd El-Latif,
\newblock ``A secure federated learning framework for {5G} networks,''
\newblock {\em IEEE Wireless Communications}, vol. 27, no. 4, pp. 24--31, 2020.

\bibitem{frey2021towards-tit}
M.~Frey, I.~Bjelaković, and S.~Stańczak,
\newblock ``Towards secure over-the-air computation,''
\newblock {\em Submitted to IEEE Transactions on Information Theory}, 2022,
\newblock Preprint available at arXiv:2001.03174.

\bibitem{frey2021towards-isit}
M.~Frey, I.~Bjelaković, and S.~Stańczak,
\newblock ``Towards secure over-the-air computation,''
\newblock in {\em 2021 IEEE International Symposium on Information Theory
  (ISIT)}. IEEE, 2021, pp. 700--705.

\bibitem{Hu2022}
C.~Hu, Q.~Li, Q.~Zhang, and J.~Qin,
\newblock ``Secure transceiver design and power control for over-the-air
  computation networks,''
\newblock {\em IEEE Communications Letters}, vol. 26, no. 7, pp. 1509--1513,
  2022.

\bibitem{Amiri2020}
M~.M. Amiri and D.~Gündüz,
\newblock ``Machine learning at the wireless edge: Distributed stochastic
  gradient descent over-the-air,''
\newblock {\em IEEE Transactions on Signal Processing}, vol. 68, pp.
  2155--2169, 2020.

\bibitem{Amiri2020_2}
M.~M. Amiri and D.~Gündüz,
\newblock ``Federated learning over wireless fading channels,''
\newblock {\em IEEE Transactions on Wireless Communications}, vol. 19, no. 5,
  pp. 3546--3557, 2020.

\bibitem{Engl1996}
H.~W. Engl, M.~Hanke, and A.~Neubauer,
\newblock {\em Regularization of Inverse Problems},
\newblock Kluwer Academic Publishers, Dordrecht, 1996.

\bibitem{Golub1996}
G.~H. Golub and C.~F. {Van Loan},
\newblock {\em Matrix Computations},
\newblock The Johns Hopkins University Press, Baltimore, 1996.

\bibitem{Abolghasemi2010}
V.~Abolghasemi, S.~Ferdowsi, B.~Makkiabadi, and S.~Sanei,
\newblock ``On optimization of the measurement matrix for compressive
  sensing,''
\newblock in {\em 2010 18th European Signal Processing Conference}, 2010, pp.
  427--431.

\bibitem{Lecun1998}
Y.~LeCun,
\newblock ``The {MNIST} database of handwritten digits,''
\newblock {\em http://yann. lecun. com/exdb/mnist/}, 1998.

\bibitem{sklearn2022}
Scikit Learn,
\newblock ``sklearn. linearmodel. lasso,''
\newblock {\em Scikit Learn,[Online]. Available: https://scikitlearn.
  org/stable/modules/generated/sklearn. linear\_model. Lasso. html.[Accessed in
  21 05 2023]}, 2022.

\bibitem{Horn1991}
R.~A. Horn and C.~R. Johnson,
\newblock {\em Topics in Matrix Analysis},
\newblock Cambridge University Press, New York, 1991.

\bibitem{Rudelson2010}
M.~Rudelson and R.~Vershynin,
\newblock ``Non-asymptotic theory of random matrices: extreme singular
  values,''
\newblock in {\em Proceedings of the International Congress of Mathematicians},
  2010.

\bibitem{Loyka2015}
S.~Loyka,
\newblock ``On singular value inequalities for the sum of two matrices,'' 2015,
\newblock arXiv.

\end{thebibliography}

\end{document}